\documentclass[pmlr,twocolumn,10pt]{jmlr} 

\usepackage{booktabs}
\usepackage{siunitx}
\usepackage{multirow}

\usepackage[switch]{lineno}

\jmlrvolume{287}
\jmlryear{2025}
\jmlrsubmitted{} 
\jmlrpublished{} 
\jmlrworkshop{Conference on Health, Inference, and Learning (CHIL) 2025}

\title[Transformer Model for Longitudinal AD Prediction]
{Transformer Model for Alzheimer's Disease Progression Prediction Using 
Longitudinal Visit Sequences}

\author{%
 \Name{Mahdi Moghaddami} \Email{moghaddami@oakland.edu}\\
 \Name{Clayton Schubring} \Email{c.schubring@oakland.edu}\\
 \Name{Mohammad-Reza Siadat} \Email{siadat@oakland.edu}\\
 \addr Oakland University, USA
}

\begin{document}

\maketitle

\begin{abstract}
    Alzheimer's disease (AD) is a neurodegenerative disorder with no known cure that
    affects tens of millions of people worldwide. Early detection of AD is critical
    for timely intervention to halt or slow the progression of the disease. In this
    study, we propose a Transformer model for predicting the stage of AD progression
    at a subject's next clinical visit using features from a sequence of visits
    extracted from the subject's visit history. We also rigorously compare our model
    to recurrent neural networks (RNNs) such as long short-term memory (LSTM), gated
    recurrent unit (GRU), and minimalRNN and assess their performances based on
    factors such as the length of prior visits and data imbalance. We test the
    importance of different feature categories and visit history, as well as compare
    the model to a newer Transformer-based model optimized for time series. Our model
    demonstrates strong predictive performance despite missing visits and missing
    features in available visits, particularly in identifying converter
    subjects—individuals transitioning to more severe disease stages—an area that
    has posed significant challenges in longitudinal prediction. The results highlight
    the model's potential in enhancing early diagnosis and patient outcomes.
\end{abstract}

\paragraph*{Data and Code Availability}
We use data provided by the Alzheimer's disease prediction of longitudinal
evolution (TADPOLE) challenge \citep{marinescu2018tadpolechallengepredictionlongitudinal}
which is derived from the Alzheimer's Disease Neuroimaging Initiative
(ADNI)
database \citep{Jack2008TheAD} and is available publicly at
\url{https://ida.loni.usc.edu/}. Our code is also available and included as
supplemental material.

\paragraph*{Institutional Review Board (IRB)}
This research does not require IRB approval.
\section{Introduction}
\label{sec:introduction}

Alzheimer's disease (AD) is the most prevalent cause of dementia
\citep{SCHELTENS2016505}.
Nearly 7 million Americans age 65 and older are living with AD, and it is
projected that by 2050, this number may grow to a projected 12.7 million
as the size of the U.S. population age 65 and older continues to grow.
AD was the fifth-leading cause of death among individuals aged 65 and
older in 2021 in the United States \citep{20242024AD}. Even
before an individual with AD dies, their quality of life and their
family's deteriorate substantially. The significant growth rate of AD,
combined with its fatality, makes AD a vital public safety matter.

Since no cure has been found for AD yet, it must be detected as early
as possible so that medical professionals can offer effective treatment
measures and lifestyle changes to reduce the speed of its progression
\citep{lancet}.
However, most diagnoses of Alzheimer's disease happen in the moderate
to late stages, indicating that the optimal time for treatment has
often already elapsed \citep{Hoang2023VisionTF}.

Mild cognitive impairment (MCI) is a condition that exists between normal
cognitive function and Alzheimer's disease (AD). It involves a slight
decline in cognitive abilities and memory, such that the patients or
their nearest family members notice some symptoms. However, individuals
at this stage do not reach the level of dementia. MCI is considered a
precursor to AD since 10-15\% of individuals with MCI will develop AD
every year \citep{Sarakhsi2022DeepLF}.

It is crucial to develop methods that can predict the onset of AD by
forecasting the trajectory of disease progression and predicting
when cognitively normal (CN) individuals will develop MCI, and when individuals
with MCI will develop AD.

With recent advances in machine learning and deep learning, researchers
have increasingly used computer-aided diagnosis (CAD) systems, primarily
various deep learning architectures, to detect AD. Most previous work has
focused on classifying subjects using a variety of data modalities into
clinical diagnoses (CN, MCI, and AD)
\citep{Alp2024JointTA,Ji2019EarlyDO,Ahmed2019EnsemblesOP,Bi2019FunctionalBN}.

Some studies categorize MCI subjects into two classes: stable MCI
(sMCI) and progressive MCI (pMCI). Subjects that do not convert to
AD in a period of time (usually three years) are labeled sMCI, and
subjects that do are labeled pMCI
\citep{Li2012DiscriminantAO,Cui2019HippocampusAB, Spasov2018APD}.
The issue with this approach is that the conversion window is
arbitrary, and there is no difference between subjects who convert
relatively early in the given time frame and subjects that convert
much later.

Using multiple time points for each subject would be preferable, as the model
would have the information to understand the subject's progression trajectory
comprehensively.
\citet{Zhang2012PredictingFC} aims to forecast future clinical changes in
patients with MCI accurately. The authors utilize a multi-kernel support
vector machine (SVM) to predict future cognitive scores at a 24-month
follow-up and whether a patient will convert from MCI to AD, using data
from time points at least 6 months before the actual conversion.
\citet{Cui2019RNNbasedLA} presents a combination of a convolutional
neural network (CNN) and an RNN to distinguish AD patients from normal
controls and classify sMCI patients vs. pMCI patients.
\citet{Nguyen2019PredictingAD} used longitudinal data from the TADPOLE
challenge and a minimalRNN model
\citep{chen2018minimalrnninterpretabletrainablerecurrent} to predict clinical
diagnosis and other biomarkers for each subject up to 6 years into the future.

\citet{Vaswani2017AttentionIA} introduced the Transformer architecture in 2017.
The authors used a mechanism called attention to transfer context between the encoder
and decoder layers in the model.
Transformer-based models have been employed in CAD systems more frequently
after yielding superior results in natural language processing.
\citet{chen2023longformerlongitudinaltransformeralzheimers}
uses a Transformer-based model to classify CN
vs. AD using magnetic resonance imaging (MRI) data from multiple visits.
\citet{Alp2024JointTA} uses a vision Transformer (ViT) to extract spatial
features from T1-weighted brain MRIs. Sequence modeling is performed on
the feature sequences to maintain interdependencies, and finally, features
are classified using a Time Series Transformer (TST)
\citep{10.1145/3447548.3467401}.

The TST is designed for time series forecasting and representation learning.
It's an encoder-only Transformer incorporating
time-series-specific inductive biases, such as causal attention and positional
encoding tailored for temporal data.

Our work differs from prior work in two main ways. Most previous works either
do not incorporate the longitudinal history of subjects or do not predict
the clinical diagnosis in the near future. Also, to the best of our knowledge,
no study focuses on assessing the significance of the length of visit history
of subjects utilizing Transformers and RNN-based models.

In this work, we propose a Transformer model that utilizes
a subject's entire visit history to predict the clinical diagnosis of the
subject at the next visit. We use an imputation strategy to handle missing
data and avoid discarding visits with missing feature values.
We focus on the conversion points of subjects
from CN to MCI and MCI to AD since they are vital for early detection.
Statistical testing is performed to compare the performance of the models
and to show that the proposed Transformer model significantly outperforms
the RNN-based models in predicting the conversion points.
Finally, we analyze feature importance and the importance of visit history
length.

\section{Materials and Methods}
\label{sec:methods}

\subsection{Data}
\label{sec:data}

We use data from the Alzheimer's disease prediction of
longitudinal evolution (TADPOLE) challenge
\footnote{\url{https://tadpole.grand-challenge.org/}}
, which contains longitudinal
data from 1677 participants
\citep{marinescu2018tadpolechallengepredictionlongitudinal} from the Alzheimer's
Disease Neuroimaging Initiative (ADNI) \citep{Mueller2005WaysTA}. Multiple data
points were collected for each participant at a minimum interval of 6 months,
making it a valuable dataset for longitudinal studies.
The TADPOLE challenge aimed to develop models to
accurately predict future clinical states and biomarker trajectories in
individuals at risk for or diagnosed with AD.

We use a subset of 23 features from the TADPOLE dataset, the same ones
recommended by the TADPOLE challenge. These features include information from
different data modalities, such as neuropsychological test scores,
clinical diagnosis (CN, MCI, or AD), MRI measures, positron emission tomography
(PET) measures, and cerebrospinal fluid (CSF) markers. \tableref{tab:features}
shows a comprehensive list of the features used in this study.

\begin{table*}[hbtp]
    \floatconts
    {tab:features}
    {\caption{List of features with their means, standard deviations, and the percentage of missing values.
            SB: Sum of boxes, ADAS: Alzheimer's Disease
            Assessment Scale, RAVLT: Rey Auditory Verbal Learning Test.}}
    {\begin{tabular}{llcc}
            \toprule
            \bfseries Category & \bfseries Feature                         & \bfseries Mean $\pm$ std   & \bfseries Missingness \% \\
            \midrule
            -                  & Diagnosis                                 & -                          & 30.11\%                  \\
            \midrule
            \multirow{10}{*}{Cognitive}
                               & Functional Activities Questionnaire (FAQ) & 4.65 $\pm$ 6.96            & 29.40\%                  \\
                               & Clinical Dementia Rating Scale (SB)       & 1.83 $\pm$ 2.29            & 29.64\%                  \\
                               & Mini-Mental State Examination (MMSE)      & 26.92 $\pm$ 3.5            & 29.88\%                  \\
                               & ADAS-Cog11                                & 10.74 $\pm$ 7.76           & 30.05\%                  \\
                               & RAVLT immediate                           & 35.08 $\pm$ 13.24          & 30.67\%                  \\
                               & RAVLT learning                            & 4.14 $\pm$ 2.81            & 30.67\%                  \\
                               & ADAS-Cog13                                & 16.62 $\pm$ 10.68          & 30.72\%                  \\
                               & RAVLT forgetting                          & 4.26 $\pm$ 2.55            & 30.88\%                  \\
                               & RAVLT forgetting percent                  & 58.73 $\pm$ 37.57          & 31.43\%                  \\
                               & Montreal Cognitive Assessment (MOCA)      & 23.52 $\pm$ 4.18           & 61.01\%                  \\
            \midrule
            \multirow{7}{*}{MRI}
                               & Intracranial volume                       & 1534699.07 $\pm$ 164732.93 & 37.57\%                  \\
                               & Whole brain volume                        & 1010781.21 $\pm$ 111280.94 & 39.65\%                  \\
                               & Ventricles volume                         & 42119.98 $\pm$ 23274.12    & 41.56\%                  \\
                               & Hippocampus volume                        & 6684.54 $\pm$ 1224.13      & 46.61\%                  \\
                               & Entorhinal cortical volume                & 3455.9 $\pm$ 801.46        & 49.22\%                  \\
                               & Fusiform cortical volume                  & 17117.41 $\pm$ 2798.63     & 49.22\%                  \\
                               & Middle temporal cortical volume           & 19206.76 $\pm$ 3098.07     & 49.22\%                  \\
            \midrule
            \multirow{5}{*}{Biomarker}
                               & Fluorodeoxyglucose (FDG) - PET            & 1.21 $\pm$ 0.16            & 73.69\%                  \\
                               & Phosphorylated tau                        & 27.59 $\pm$ 11.7           & 81.38\%                  \\
                               & Beta-amyloid (CSF)                        & 1052.48 $\pm$ 502.57       & 81.40\%                  \\
                               & Total tau                                 & 288.67 $\pm$ 105.95        & 81.45\%                  \\
                               & Florbetapir (18F-AV-45) - PET             & 1.19 $\pm$ 0.22            & 83.38\%                  \\
            \bottomrule
        \end{tabular}}
\end{table*}

\begin{figure*}[htbp]
    \floatconts
    {fig:distibution}
    {\caption{Distribution of visit sequences by group number}}
    {%
        \subfigure[Raw dataset]{\label{fig:dist_raw}
            \includegraphics[width=0.468\linewidth]{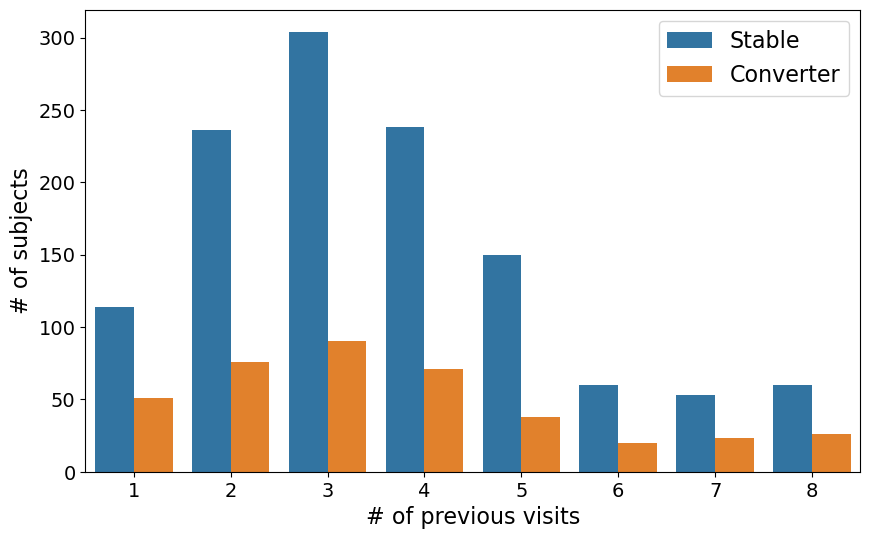}}
        \qquad
        \subfigure[Balanced dataset]{\label{fig:fist_balanced}
            \includegraphics[width=0.468\linewidth]{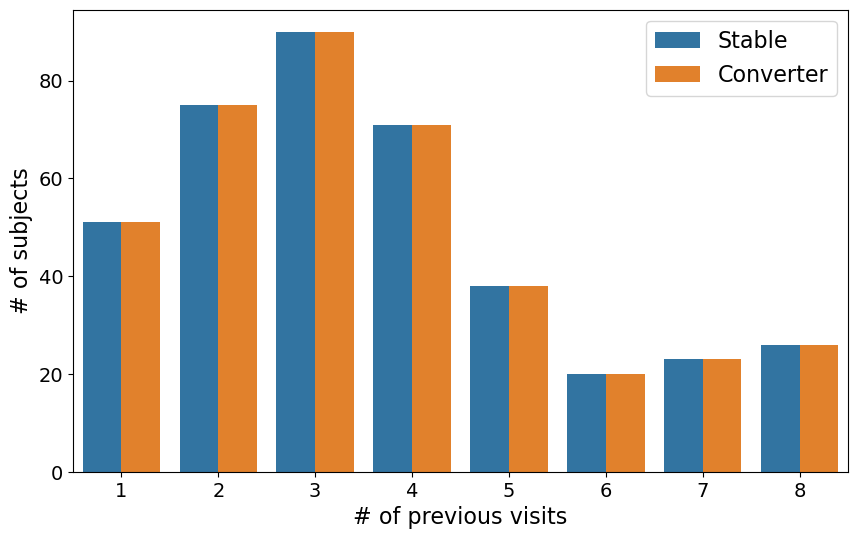}}
    }
\end{figure*}   

\subsection{Preprocessing}
\label{sec:preprocessing}
Like many long-term studies, the TADPOLE dataset suffers from some problems:
follow-up appointments not occurring, irregular scheduling of visits, and
some subjects dropping out before the completion of the study. Also, some
measurements are not taken at every visit due to the procedure's invasiveness.
These issues, combined with unintentional mistakes during data collection or scanning
cause missing (See \tableref{tab:features}) and incorrect values in the dataset.

To address these points, we perform the following preprocessing steps:
\begin{enumerate}
    \item Drop all visits belonging to subjects with only one visit.
    \item Drop all visits belonging to \textit{reverter} subjects. These are
          subjects who convert from AD to MCI or from MCI to CN at any point in
          their visit history.
    \item There are subjects who convert from CN to MCI or from MCI to AD multiple
          times. We identify these subjects and discard all their visits dated
          after the first conversion.
    \item Drop all visits with a missing value for clinical diagnosis.
\end{enumerate}


To handle missing data, we use an imputation method called ``model filling''
proposed by \citet{Nguyen2019PredictingAD}. This method uses predictions from
an RNN model to fill in missing values in the dataset. We use predictions from
a minimalRNN model.

Since there are irregular time gaps between visits, we introduce an additional
feature to encode temporal information into every visit in a subject's visit
history. This feature represents the number of months until the final visit.

\subsection{Sequence Generation and Dataset Construction}
\label{sec:sequence_generation}

To better focus on the key points of conversion in participants' visit history, we
construct datasets using stable and converter sequences. We define these terms
as follows:

\begin{definition}[Converter Sequence]
    \label{def:converter_sequence}
    A sequence of visits for a subject sorted by examination date where at
    the last visit in the sequence, the subject converts to the next stage
    of the disease (CN to MCI, CN to AD, or MCI to AD).
\end{definition}

\begin{definition}[Stable Sequence]
    \label{def:stable_sequence}
    A sequence of visits for a subject sorted by examination date where at the
    last visit in the sequence, the subject's diagnosis remains the same as the
    penultimate visit (CN to CN, MCI to MCI, or AD to AD).
\end{definition}

We generate one converter sequence of maximal length for every subject who at
any point in their visit history, converted from CN to MCI or from MCI to AD.
Similarly, we generate one stable sequence of maximal length from every other
subject. The set of all the stable and converter sequences we obtain by this
process is the \textit{raw} dataset. The raw dataset is highly imbalanced,
containing 1222 stable sequences and only 405 converter sequences, of which
only 70 contain a CN to MCI conversion.

\begin{definition}[Group $n$]
    \label{def:group_n}
    All stable or converter sequences with $n+1$ visits.
\end{definition}

Note that every sequence only belongs to one group and groups cannot share
any sequences.

We also construct one additional dataset: the \textit{balanced} dataset.
This dataset contains an equal number of stable and
converter sequences in each group. It is created in order to ensure that
the models learn equally from all classes, preventing bias towards the
majority class. This dataset is built by randomly discarding stable
sequences in each group until the number of stable and converter sequences
in each group is equal.


\figureref{fig:distibution} shows the distribution of visit sequences by
group number in the raw and balanced datasets. \figureref{fig:future_dx_distibution}
shows the distribution of the diagnosis (DX) in the target visit by group number
in the raw dataset.

\begin{figure}[htpb]
    \floatconts
    {fig:future_dx_distibution}
    {\caption{Distribution of Future DX across groups}}
    {\includegraphics[width=\linewidth]{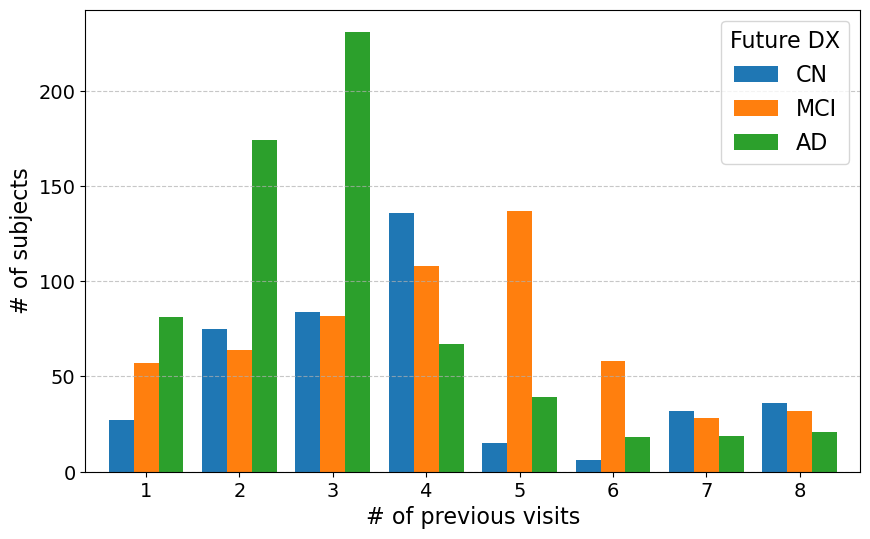}}
\end{figure}

\subsection{Model}
\label{sec:model}
We consider a multi-class classification (CN, MCI, AD) problem with the input being
the first $n$ visits from a visit sequence $v=(v_1, v_2, \ldots, v_n)$ and the
target being the diagnosis at visit $n+1$ ($d_{n+1}$).

We utilized three RNN-based models as a baseline for predicting disease
progression: long short-term memory (LSTM) \citep{hochreiter1997long},
gated recurrent unit (GRU) \citep{cho2014learning},
and minimalRNN (minRNN) \citep{chen2018minimalrnninterpretabletrainablerecurrent}.
RNN-based models try to preserve the context
of previous time points by carrying a hidden state vector $h$. At time point $t$,
predictions for the next time point ($\hat{y_t}$) are made using the current
observation $x_t$ and the hidden state from the prior time point $h_{t-1}$. The
equations for updating the hidden state and calculating the output vary between
LSTM, GRU, and minRNN.

\begin{figure}[htbp]
    \floatconts
    {fig:rnn}
    {\caption{Prediction process for RNN-based models}}
    {\includegraphics[width=\linewidth]{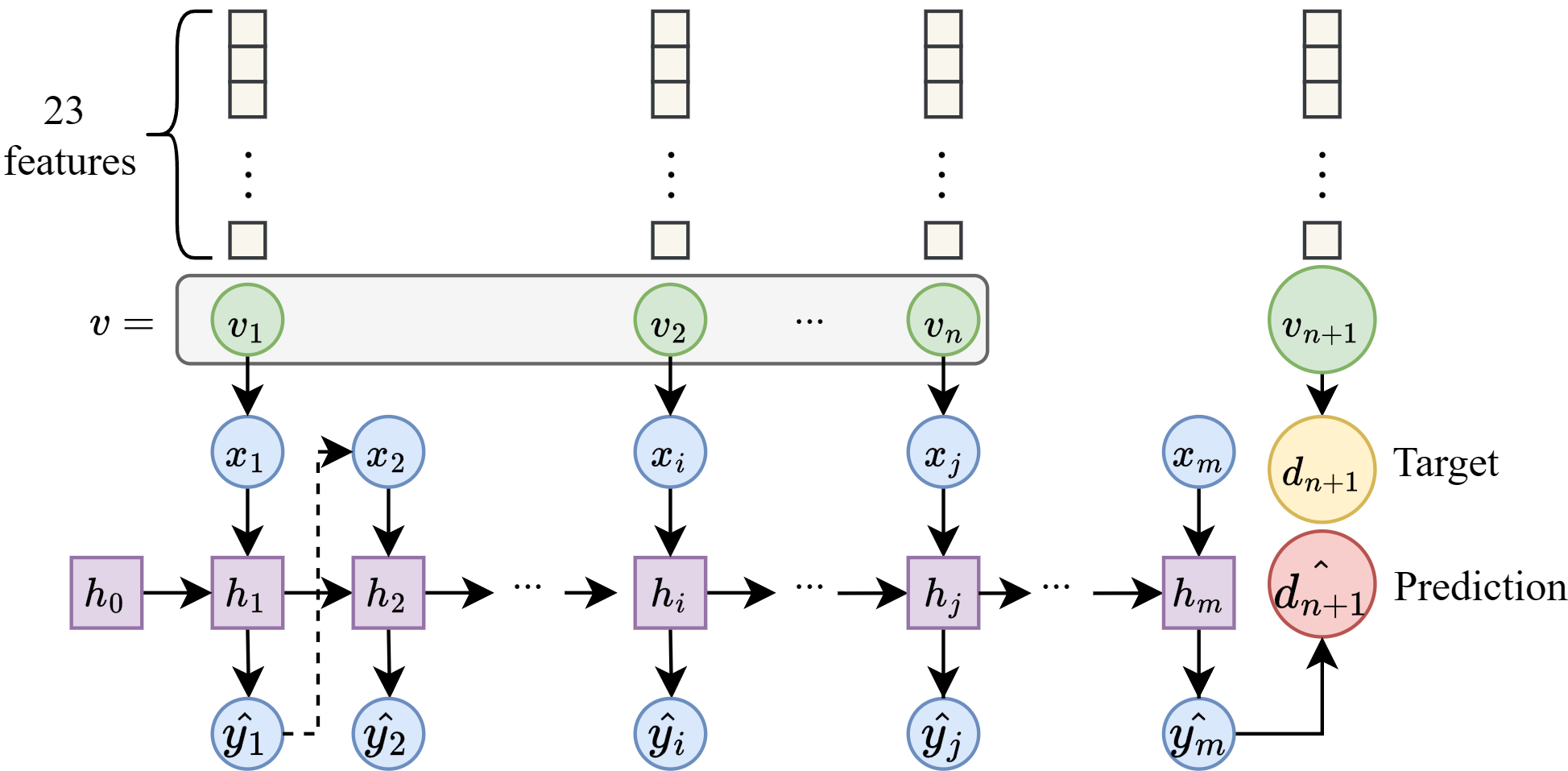}}
\end{figure}

Following \citet{Nguyen2019PredictingAD}, we train these models by predicting
clinical diagnosis
for every month into the future up to six years, starting with only the first
visit in the sequence $v_1$. If the actual feature values for timepoint $t$ exist in the
visit sequence $v$, a loss value is calculated between $\hat{y_{t-1}}$ and $x_t$ and
weights are updated accordingly; otherwise, $\hat{y_{t-1}}$ will be used as the input
($x_t$) for the next month. \figureref{fig:rnn} shows an overview of the
training and prediction process for RNN-based models. They continuously make predictions
for the next month until the end of the visit sequence is reached. At this point,
the model's prediction for the diagnosis in the last visit is compared to the actual
diagnosis.

Our proposed model is based on the Transformer architecture
\citep{Vaswani2017AttentionIA}.
Unlike an RNN-based model, the Transformer can take in the entire input visit
sequence at once, increasing training speed significantly. Each visit in an
input visit sequence represents a token to the Transformer. The clinical
diagnosis of the given sequence in the next visit ($d_{n+1}$), given their visit
sequence $v$, is calculated by the Transformer as depicted in \figureref{fig:Transformer}.

\begin{figure}[htbp]
    \floatconts
    {fig:Transformer}
    {\caption{Prediction process for the Transformer}}
    {\includegraphics[width=\linewidth]{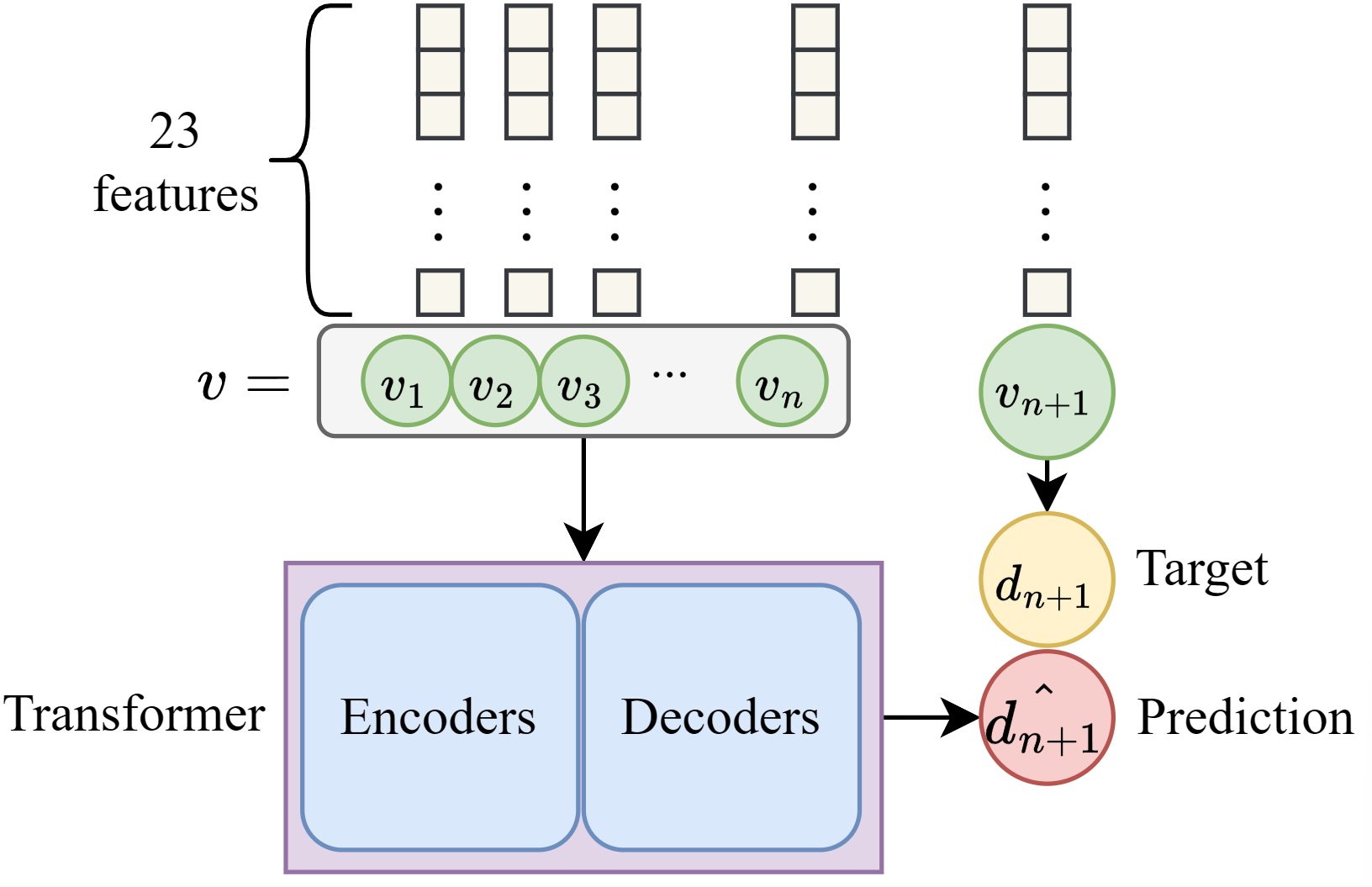}}
\end{figure}

To compare our proposed model to a newer architecture, we also consider the
Time Series Transformer (TST) \citep{10.1145/3447548.3467401}.

All models are implemented using the PyTorch Python Library
\citep{paszke2019pytorch}. Some hyperparameter
values worth mentioning are the number of encoder layers (4), the number of
decoder layers (8), the number of attention heads (4), and input dimensionality (256).
The total number of parameters for the Transformer model is 6,603,267.


\subsection{Training and Evaluation}
\label{sec:training}

We train and evaluate separate models on raw and balanced datasets.
We use 10-fold cross-validation to train and evaluate the models. Each fold is
divided such that 10\% of each group will be in the test split and 90\%
of each group will be in the train split. We split the data temporally, ensuring
that each fold has an equal proportion of sequences from each group.

We use grid search to tune appropriate hyperparameter values (number of encoder/decoder
layers, number of attention heads, dimensionality of the input layer/classification
layer).

To evaluate model performance, we compute the multiclass area under the curve
(mAUC) following \citet{Hand2001ASG}, along with the F1 score. We also report raw
accuracy, sensitivity, and specificity for each of the three classes. For each
class \( C_i \) (CN, MCI, AD), sensitivity  and specificity are defined as:
\begin{equation*}
    \text{Sensitivity}(C_i) = \frac{TP_i}{TP_i + FN_i}\,,
\end{equation*}

\begin{equation*}
    \text{Specificity}(C_i) = \frac{TN_i}{TN_i + FP_i} \,,
\end{equation*}

where \( TP_i \), \( FN_i \), \( TN_i \), and \( FP_i \) represent the number of
true positives, false negatives, true negatives, and false positives, respectively,
for class \( C_i \). The BCA is then calculated by taking the mean of the sensitivity
(Sens.) and specificity (Spec.) for each class, averaged over all 3 classes:
\begin{equation*}
    \text{BCA} = \frac{1}{3} \sum_{i=1}^3 \left(\frac{1}{2} \bigg( \text{Sens.}(C_i) + \text{Spec.}(C_i) \bigg) \right)\,.
\end{equation*}

In addition to evaluating these metrics on the full dataset, we compute them
separately for two subsets of the data: one containing only stable sequences
and the other containing only converter sequences.

\section{Results}
\label{sec:results}

All statistical comparisons were performed using a two-sample Welch's t-test
with $n_1 = 10$ and $n_2 = 10$ for each model. The results are reported as mean
$\pm$ standard deviation across the ten folds.

\subsection{Raw Dataset Results}

On the raw dataset, the TST recorded the highest mAUC, while the Transformer had the
highest F1 score (\tableref{tab:maucraw}). The Transformer model demonstrated the
highest overall accuracy compared to the baseline RNN models. BCA followed the same
trend, with the Transformer outperforming GRU, LSTM, and minRNN. Additionally, the
Transformer exhibited the highest sensitivity and specificity.

\begin{table}[htbp]
    \floatconts
    {tab:maucraw}
    {\caption{mAUC and F1 score (raw dataset)}}
    {\begin{tabular}{lccc}
            \toprule
            \bfseries Model & \bfseries mAUC            & \bfseries F1 score        \\
            \midrule
            TST             & \textbf{0.938 $\pm$ 0.01} & 0.812 $\pm$ 0.03          \\
            Transformer     & 0.920 $\pm$ 0.02          & \textbf{0.824 $\pm$ 0.04} \\
            GRU             & 0.846 $\pm$ 0.11          & 0.786 $\pm$ 0.04          \\
            LSTM            & 0.857 $\pm$ 0.11          & 0.786 $\pm$ 0.05          \\
            minRNN          & 0.873 $\pm$ 0.11          & 0.766 $\pm$ 0.05          \\
            \bottomrule
        \end{tabular}}
\end{table}

Statistical testing revealed that the Transformer significantly outperformed all
RNN models in BCA, even the best RNN (GRU) with \textit{p} = 4.12e-02. This came
from a statistically significant difference in sensitivity (\textit{p} = 3.85e-02),
while the difference in specificity was not significant. With regards to raw accuracy,
the Transformer outperformed minRNN (\textit{p} = 1.53e-02) but otherwise, the
Transformer-based models were not significantly different than the RNNs.

\begin{table*}[!ht]
    \floatconts
    {tab:raw_performance}
    {\caption{Performance metrics for the raw dataset across all subsets}}
    {\begin{tabular}{lcccccc}
            \toprule
            \bfseries Subset & \bfseries Metric & \bfseries  TST              & \bfseries Transformer     & \bfseries GRU             & \bfseries LSTM   & \bfseries  minRNN         \\
            \midrule
            \multirow{4}{*}{Overall}
                             & Acc.             & 0.814 $\pm$ 0.04            & \textbf{0.824 $\pm$ 0.04} & 0.792 $\pm$ 0.04          & 0.789 $\pm$ 0.04 & 0.777 $\pm$ 0.04          \\
                             & BCA              & 0.866 $\pm$ 0.02            & \textbf{0.874 $\pm$ 0.03} & 0.830 $\pm$ 0.05          & 0.829 $\pm$ 0.06 & 0.819 $\pm$ 0.06          \\
                             & Sens.            & 0.828 $\pm$0.03             & \textbf{0.837 $\pm$ 0.04} & 0.777 $\pm$ 0.08          & 0.773 $\pm$ 0.08 & 0.761 $\pm$ 0.09          \\
                             & Spec.            & 0.904 $\pm$ 0.02            & \textbf{0.910 $\pm$ 0.02} & 0.884 $\pm$ 0.04          & 0.885 $\pm$ 0.04 & 0.877 $\pm$ 0.04          \\
            \midrule
            \multirow{4}{*}{Stable}
                             & Acc.             & 0.856 $\pm$ 0.04            & 0.883 $\pm$ 0.04          & 0.984 $\pm$ 0.01          & 0.973 $\pm$ 0.03 & \textbf{0.986 $\pm$ 0.01} \\
                             & BCA              & 0.907  $\pm$ 0.02           & 0.922 $\pm$ 0.02          & \textbf{0.951 $\pm$ 0.05} & 0.939 $\pm$ 0.07 & 0.950 $\pm$ 0.05          \\
                             & Sens.            & 0.880  $\pm$ 0.03           & 0.898 $\pm$ 0.03          & \textbf{0.921 $\pm$ 0.08} & 0.905 $\pm$ 0.10 & 0.920 $\pm$ 0.08          \\
                             & Spec.            & 0.934 $\pm$ 0.02            & 0.945 $\pm$ 0.02          & \textbf{0.981 $\pm$ 0.03} & 0.972 $\pm$ 0.04 & \textbf{0.981 $\pm$ 0.03} \\
            \midrule
            \multirow{4}{*}{Converter}
                             & Acc.             & \textbf{0.694  $\pm$ 0.08}  & 0.648 $\pm$ 0.14          & 0.184 $\pm$ 0.07          & 0.217 $\pm$ 0.08 & 0.129 $\pm$ 0.07          \\
                             & BCA              & \textbf{0.662   $\pm$ 0.04} & 0.648 $\pm$ 0.05          & 0.464 $\pm$ 0.03          & 0.476 $\pm$ 0.04 & 0.442 $\pm$ 0.03          \\
                             & Sens.            & \textbf{0.446  $\pm$ 0.03}  & 0.430 $\pm$ 0.05          & 0.320 $\pm$ 0.04          & 0.334 $\pm$ 0.05 & 0.298 $\pm$ 0.04          \\
                             & Spec.            & \textbf{0.879  $\pm$ 0.05}  & 0.866 $\pm$ 0.06          & 0.609 $\pm$ 0.02          & 0.618 $\pm$ 0.03 & 0.586 $\pm$ 0.03          \\
            \bottomrule
        \end{tabular}
    }
\end{table*}

\begin{table*}[!htbp]
    \floatconts
    {tab:balanced_performance}
    {\caption{Performance metrics for the balanced dataset across all subsets}}
    {\begin{tabular}{lcccccc}
            \toprule
            \bfseries Subset & \bfseries Metric & \bfseries TST       & \bfseries Transformer     & \bfseries GRU             & \bfseries LSTM   & \bfseries minRNN \\
            \midrule
            \multirow{4}{*}{Overall}
                             & Acc.             & 0.779 $\pm$  0.04   & \textbf{0.802 $\pm$ 0.05} & 0.648 $\pm$ 0.04          & 0.684 $\pm$ 0.07 & 0.623 $\pm$ 0.04 \\
                             & BCA              & 0.822 $\pm$  0.03   & \textbf{0.839 $\pm$ 0.04} & 0.758 $\pm$ 0.03          & 0.770 $\pm$ 0.05 & 0.742 $\pm$ 0.03 \\
                             & Sens.            & 0.766 $\pm$ 0.04    & \textbf{0.788 $\pm$ 0.05} & 0.684 $\pm$ 0.03          & 0.699 $\pm$ 0.06 & 0.666 $\pm$ 0.04 \\
                             & Spec.            & 0.878 $\pm$ 0.03    & \textbf{0.890 $\pm$ 0.03} & 0.832 $\pm$ 0.02          & 0.842 $\pm$ 0.04 & 0.818 $\pm$ 0.03 \\
            \midrule
            \multirow{4}{*}{Stable}
                             & Acc.             & 0.787 $\pm$ 0.07    & 0.814 $\pm$ 0.07          & \textbf{0.974 $\pm$ 0.02} & 0.926 $\pm$ 0.06 & 0.956 $\pm$ 0.03 \\
                             & BCA              & 0.866   $\pm$  0.03 & 0.883 $\pm$ 0.05          & \textbf{0.910 $\pm$ 0.07} & 0.885 $\pm$ 0.08 & 0.900 $\pm$ 0.06 \\
                             & Sens.            & 0.829 $\pm$ 0.04    & 0.849 $\pm$ 0.06          & \textbf{0.859 $\pm$ 0.09} & 0.832 $\pm$ 0.11 & 0.849 $\pm$ 0.08 \\
                             & Spec.            & 0.904  $\pm$ 0.03   & 0.917 $\pm$ 0.03          & \textbf{0.960 $\pm$ 0.05} & 0.939 $\pm$ 0.06 & 0.951 $\pm$ 0.04 \\
            \midrule
            \multirow{4}{*}{Converter}
                             & Acc.             & 0.775  $\pm$ 0.06   & \textbf{0.792 $\pm$ 0.05} & 0.323 $\pm$ 0.08          & 0.439 $\pm$ 0.14 & 0.291 $\pm$ 0.05 \\
                             & BCA              & 0.703  $\pm$  0.03  & \textbf{0.713 $\pm$ 0.03} & 0.512 $\pm$ 0.03          & 0.566 $\pm$ 0.06 & 0.499 $\pm$ 0.02 \\
                             & Sens.            & 0.495 $\pm$ 0.05    & \textbf{0.504 $\pm$ 0.05} & 0.358 $\pm$ 0.05          & 0.419 $\pm$ 0.08 & 0.346 $\pm$ 0.04 \\
                             & Spec.            & 0.910 $\pm$ 0.03    & \textbf{0.922 $\pm$ 0.02} & 0.666 $\pm$ 0.02          & 0.713 $\pm$ 0.05 & 0.651 $\pm$ 0.02 \\
            \bottomrule
        \end{tabular}}
\end{table*}

The significant difference between the two architectures is most apparent when
considering the performance on stable and converter sequences. For stable sequences,
the Transformer model underperformed relative to all RNN baselines, exhibiting lower
accuracy, BCA, sensitivity, and specificity. Statistical tests confirmed that the
Transformer was significantly worse in accuracy than GRU (\textit{p} = 7.85e-06),
LSTM (\textit{p} = 1.71e-05), and minRNN (\textit{p} = 9.38e-06). Furthermore, the
Transformer had significantly lower specificity than GRU (\textit{p} = 3.81e-03),
and minRNN (\textit{p} = 4.37e-03). No significant differences were observed in BCA
or sensitivity for this subset of the data.

The TST achieved the highest accuracy on converter sequences, significantly
outperforming all RNN models. This trend was consistent across BCA, sensitivity, and
specificity as well, with both Transformer-based models outperforming the RNNs.
Statistical analysis confirmed these results, with TST demonstrating significantly
higher accuracy (\textit{p} = 1.48e-10), BCA (\textit{p} = 1.01e-10), sensitivity
(\textit{p} = 1.13e-06), and specificity (\textit{p} = 1.46e-10) when compared to
LSTM, the highest-performing RNN for this subset.

\tableref{tab:raw_performance} includes a comprehensive list of the values for each
metric pertaining to the overall, stable, and converter performance.

\subsection{Balanced Dataset Results}

\begin{table}[!ht]
    \floatconts
    {tab:maucbalanced}
    {\caption{mAUC and F1 score (balanced dataset)}}
    {\begin{tabular}{lccc}
            \toprule
            \bfseries Model & \bfseries mAUC             & \bfseries F1 score         \\
            \midrule
            TST             & \textbf{0.890 $\pm$ 0.033} & 0.721 $\pm$ 0.045          \\
            Transformer     & 0.883 $\pm$ 0.024          & 0.746 $\pm$ 0.064          \\
            GRU             & 0.846 $\pm$ 0.107          & \textbf{0.786 $\pm$ 0.041} \\
            LSTM            & 0.857 $\pm$ 0.114          & \textbf{0.786 $\pm$ 0.048} \\
            minRNN          & 0.873 $\pm$ 0.114          & 0.766 $\pm$ 0.049          \\
            \bottomrule
        \end{tabular}}
\end{table}

On the balanced dataset, the TST recorded the highest mAUC, while the RNNs had
higher F1 scores. The Transformer model outperformed all of the baseline RNNs
across accuracy, BCA, sensitivity, and specificity. Statistical testing confirmed
that the Transformer was significantly better than all RNN models for all metrics;
for the best RNN (LSTM), we observed a significant difference in accuracy
(\textit{p} = 6.59e-04), BCA (\textit{p} = 2.44e-03), sensitivity (\textit{p} =
1.97e-03), and specificity (\textit{p} = 5.19e-03).

For stable sequences, the Transformer model exhibited lower accuracy, BCA,
sensitivity, and specificity compared to all RNN models. Statistical tests
confirmed that the Transformer was significantly worse in accuracy than GRU
(\textit{p} = 4.08e-02), LSTM (\textit{p} = 1.63e-01), and minRNN (\textit{p}
= 6.74e-02). However, the differences across models in BCA were not statistically
significant. The differences in sensitivity and specificity were also not
significant, besides that GRU was significantly better in specificity (\textit{p}
= 2.79e-02).

The Transformer achieved the highest accuracy on converter sequences,
substantially outperforming each of the RNNs. This trend was also consistent
for BCA, sensitivity, and specificity. This difference was statistically
significant; when compared to the best RNN in this subset (LSTM), we observe
significantly higher accuracy (\textit{p} = 1.20e-05), BCA (\textit{p} = 1.57e-05),
sensitivity (\textit{p} = 1.25e-02), and specificity (\textit{p} = 5.16e-08).

\tableref{tab:balanced_performance} includes a comprehensive list of the values
for each metric pertaining to the overall, stable, and converter performance.

\subsection{Visit Sequence Length Assessment}

Overall, both the Transformer and RNN models exhibit similar trends in performance
across different visit sequence lengths, as illustrated in
\figureref{fig:overallgroups}. However, a key distinction emerges when comparing
stable and converter sequences, particularly among those with limited sequence
history. The Transformer model consistently achieves higher BCA for converters
in these early visit groups. This improvement, however, comes at the expense of
lower performance in stable sequences.

Statistical analysis confirms this contrast in predictive performance. Among
sequences with shorter visit histories, the Transformer significantly outperforms
all RNN models in distinguishing converters, with \textit{p}-values reaching as
low as 4.24e-10 in the earliest group. The Transformer maintains a significant
advantage in groups 2 and 3 (\textit{p} = 1.04e-05 for group 3). Beyond this point,
differences between models become statistically insignificant. The overall
trajectory of performance for converter sequences is depicted in
\figureref{fig:convertergroups}.

This improvement in converter classification comes with a corresponding reduction
in stable sequence performance, particularly in early visit groups. The
Transformer performs significantly worse than all RNN models in these cases,
with statistical significance reaching \textit{p} = 3.41e-03 in the shortest
sequence group and remaining significant across subsequent groups. These can be
seen in \figureref{fig:stablegroups}. Outside of the earliest visit groups and
a few other exceptions, model performances are statistically indistinguishable
beyond group 4, with a few minor exceptions.

An ablation study was conducted in order to further investigate the importance of
visit history. Additional Transformer models were trained using a limited number
of visits from each sequence. Specifically, one model was trained using only the
penultimate visit (final visit prior to the target visit), while another was
trained using up to the last four visits, depending on how many were available.
The results are summarized in \tableref{tab:visit_ablation}.

\begin{figure*}[!ht]
    \floatconts
    {fig:overallgroups}
    {\caption{Model BCA on all sequences}}
    {%
        \subfigure[Raw overall results]{\label{fig:rawoverallgroups}
            \includegraphics[width=0.41\linewidth]{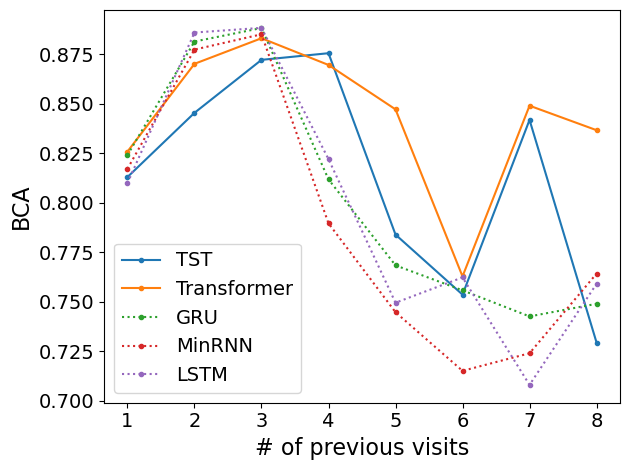}}
        \quad
        \subfigure[Balanced overall results]{\label{fig:balancedoverallgroups}
            \includegraphics[width=0.41\linewidth]{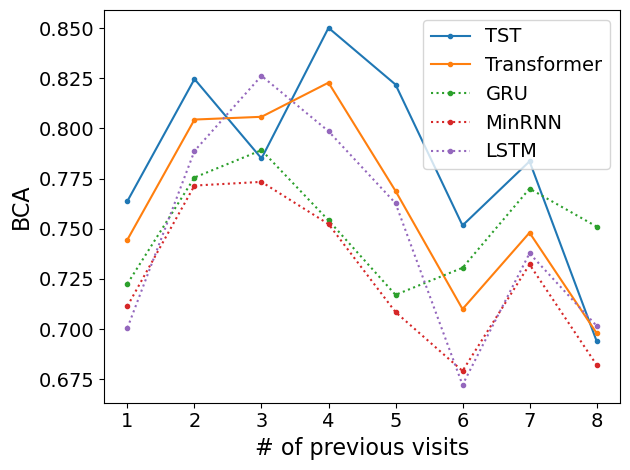}}
    }
\end{figure*}

\begin{figure*}[!ht]
    \floatconts
    {fig:stablegroups}
    {\caption{Model BCA on stable sequences}}
    {%
        \subfigure[Raw stable results]{\label{fig:rawstablegroups}
            \includegraphics[width=0.41\linewidth]{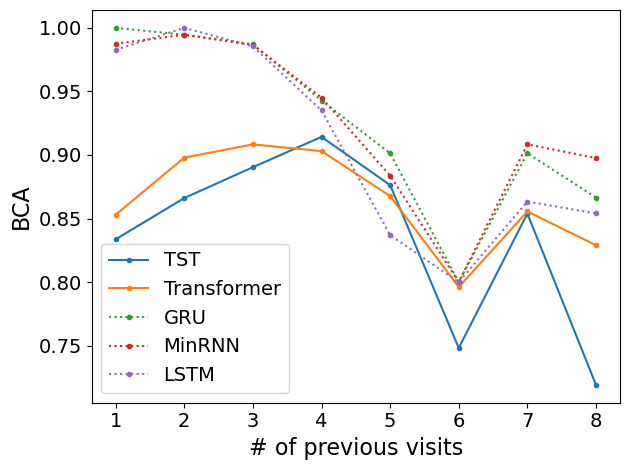}}
        \quad
        \subfigure[Balanced stable results]{\label{fig:balancedstablegroups}
            \includegraphics[width=0.41\linewidth]{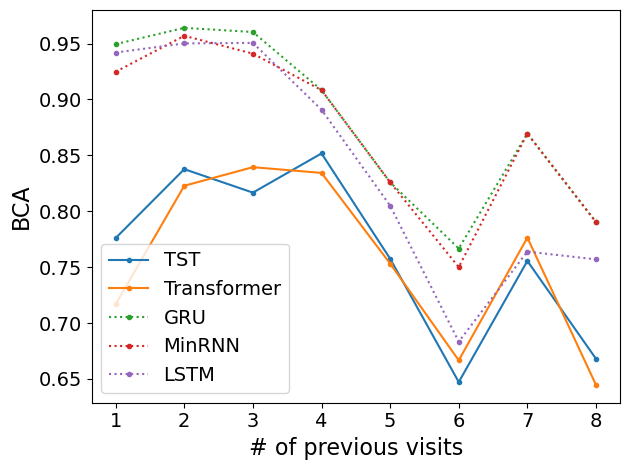}}
    }
\end{figure*}

\begin{figure*}[!ht]
    \floatconts
    {fig:convertergroups}
    {\caption{Model BCA on converter sequences}}
    {%
        \subfigure[Raw converter results]{\label{fig:rawconvertergroups}
            \includegraphics[width=0.41\linewidth]{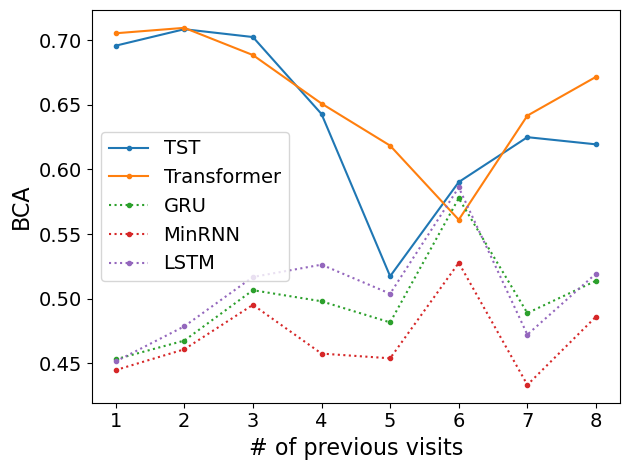}}
        \quad
        \subfigure[Balanced converter results]{\label{fig:balancedconvertergroups}
            \includegraphics[width=0.41\linewidth]{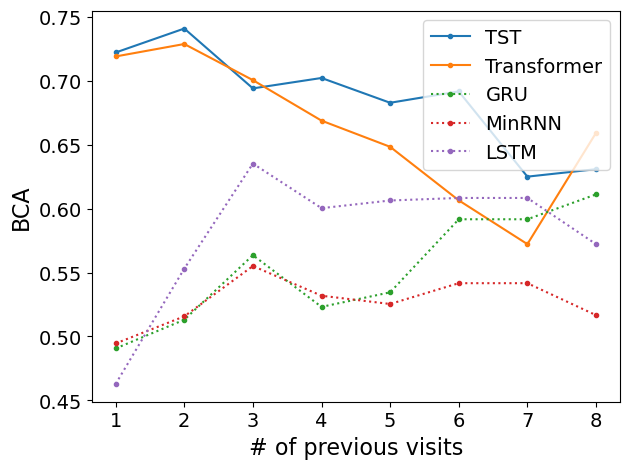}}
    }
\end{figure*}

\begin{table}[htbp]
    \floatconts
    {tab:visit_ablation}
    {\caption{Transformer BCA change (\%) when trained on limited visit history (raw dataset)}}
    {\begin{tabular}{lcc}
            \toprule
            \bfseries Visit History & \bfseries BCA & \bfseries \% Change \\
            \midrule
            Last $\leq$ 4 Visits    & 0.830         & -5.040\%            \\
            Last Visit Only         & 0.816         & -6.619\%            \\
            \bottomrule
        \end{tabular}}
\end{table}

\subsection{Feature Importance Assessment}

To assess the contribution of different feature categories to model performance,
an ablation study was conducted using the Transformer model on the raw dataset.
Categories of features were systematically removed, including cognitive scores,
volumetric MRI data, and biomarkers. Separate models were trained with these reduced
feature sets, and the resulting change in BCA was recorded. The results, contained
in \tableref{tab:feature_ablation}, illustrate the impact of each feature category
on model performance.

\begin{table}[htbp]
    \floatconts
    {tab:feature_ablation}
    {\caption{Transformer BCA change (\%) when removing a feature category (raw dataset)}}
    {\begin{tabular}{lcc}
            \toprule
            \bfseries Feature Set Removed & \bfseries BCA & \bfseries \% Change \\
            \midrule
            Cognitive Scores              & 0.736         & -15.761\%           \\
            Volumetric MRI                & 0.874         & 0.064\%             \\
            Biomarkers                    & 0.868         & -1.276\%            \\
            \bottomrule
        \end{tabular}}
\end{table}

In addition to the ablation study, separate models were trained using only cognitive
scores, only volumetric MRI data, and only biomarkers. Comparing the BCA of these
single-feature-category models to that of the full-feature model provides further
insight into the relative importance of each feature category, as shown in
\tableref{tab:feature_isolation}.

\begin{table}[htbp]
    \floatconts
    {tab:feature_isolation}
    {\caption{Transformer BCA change (\%) when trained on a single feature category (raw dataset)}}
    {\begin{tabular}{lcc}
            \toprule
            \bfseries Feature Set Used & \bfseries BCA & \bfseries \% Change \\
            \midrule
            Cognitive Scores           & 0.868         & -0.655\%            \\
            Volumetric MRI             & 0.792         & -9.314\%            \\
            Biomarkers                 & 0.821         & -6.065\%            \\
            \bottomrule
        \end{tabular}}
\end{table}
\section{Discussion}
\label{sec:discussion}

The results indicate that Transformer-based models significantly outperformed
RNNs in identifying converter sequences. In contrast, the RNN models, especially
on raw data, appeared to perform well but primarily by predicting stable sequences,
often failing to recognize converters.
While the Transformer performed slightly worse for stable sequences, this trade-off
was offset by a substantial improvement in predicting converters. Identifying
converters is critical in Alzheimer's disease research because early detection
allows for timely intervention, which can slow disease progression \citep{lancet}.
Thus, a false negative carries a much higher cost than a false positive.

Further analysis of the length of visit sequences reveals deeper insights into these
differences. Both model types followed similar performance trends across varying
sequence lengths, but key distinctions emerged in the stable and converter subsets.
The Transformer exhibited a clear advantage in predicting conversion, particularly
for shorter sequence histories. Statistical analysis confirmed that this advantage
was most pronounced in the earliest visit groups, with significant performance
differences between the Transformer-based models and all RNN baselines. This
improvement in conversion prediction came at the slight expense of stable sequence
classification, particularly in early visit groups.

Interestingly, model performance does not consistently improve for longer visit
sequences, contrary to expectations. While this trend may suggest difficulties
in generalizing to longer sequences, we propose several alternative factors that
could contribute to this pattern. First, we note that there is much less data for
these higher groups; as evident in \figureref{fig:dist_raw}, groups 5-8 only account
for about 25\% of the raw dataset. Also, as was seen in
\figureref{fig:future_dx_distibution}, the future diagnosis for groups 5 and 6 is
disproportionately composed of MCI, which is notoriously challenging to predict.
Additionally, we observed that longer visit histories tend to exhibit greater
intervals between consecutive visits (10.02 months between visits on average,
compared to 6.81), introducing more inconsistencies in the temporal structure
of the data.

\begin{figure}[htbp]
    \centering
    \includegraphics[width=\linewidth]{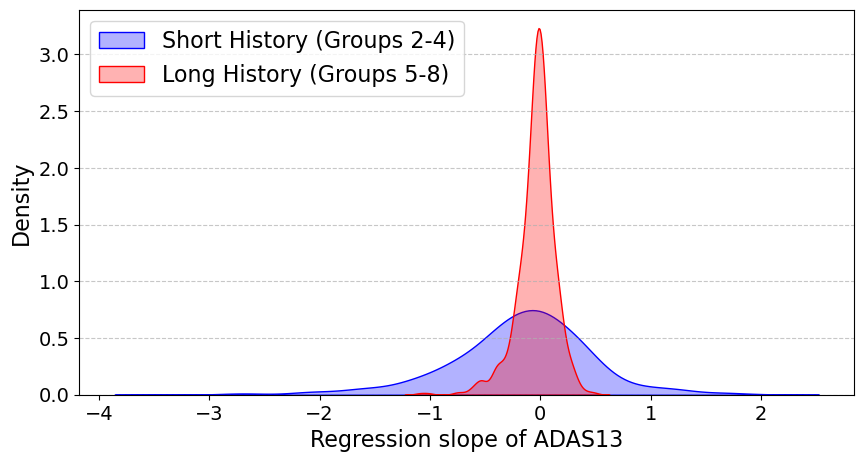}
    \caption{Regression slope of ADAS13 scores across visit sequences, short vs. long history}
    \label{fig:adas13change}
\end{figure}

Another key observation is that longer visit histories tend to exhibit a weaker
overall trend across visits. A linear regression analysis of each feature reveals
that despite having more data points, sequences in groups 5-8 generally have smaller
regression slopes, indicating less pronounced change over time.
\figureref{fig:adas13change} illustrates this effect for ADAS13 scores across the entire
raw dataset.

Despite these challenges, the model still shows evidence of learning from longer visit
histories, as seen in \tableref{tab:visit_ablation}. These findings suggest that the
model leverages information from longer visit histories, as performance degrades when
fewer visits are provided. However, the decrease in BCA is modest in both cases, which
highlights the importance of the most recent visits in making an accurate diagnosis.
This result aligns with clinical expectations, as later visits likely contain more
relevant information regarding disease progression. Nonetheless, the ability of the
full-sequence model to outperform the ablated versions reinforces the value of visit
history, even when this may introduce additional challenges.

The results from the feature ablation studies highlight the central role of cognitive
scores in predicting disease progression. The substantial change in BCA when cognitive
scores were removed (-15.76\%) suggests that they provide the most information, while
the minimal impact of removing volumetric MRI and biomarkers indicates that these
features contribute less directly to model performance. When used in isolation,
cognitive scores alone achieved nearly the same BCA as the full feature set, whereas
volumetric MRI and biomarker features yielded greater performance declines.

Further analysis of misclassified sequences reinforces the importance of cognitive
scores. Many classification errors occurred when cognitive scores were inconsistent
with the overall average for that class. Better cognitive scores often led to
incorrect predictions of stability in individuals who later converted, while low
scores sometimes resulted in the opposite. An example of the former can be seen
in \figureref{fig:adas13}, where the AD sequences that were misclassified as MCI
by the Transformer exhibited much lower ADAS13 scores in the penultimate visit
than that of the typical AD visit. The difference in these distributions is
statistically significant according to the results of a Mann-Whitney U test
(\textit{p} = 1.89e-19). We found that this is the case for many of the cognitive
scores when comparing misclassified sequences to the rest of the data. These patterns
suggest that while cognitive scores are highly predictive, the model may struggle
to account for individual variability in testing performance.

\begin{figure}[htbp]
    \floatconts
    {fig:adas13}
    {\caption{ADAS13 distribution: All AD visits vs. penultimate visits for AD sequences incorrectly classified as MCI.}}
    {\includegraphics[width=\linewidth]{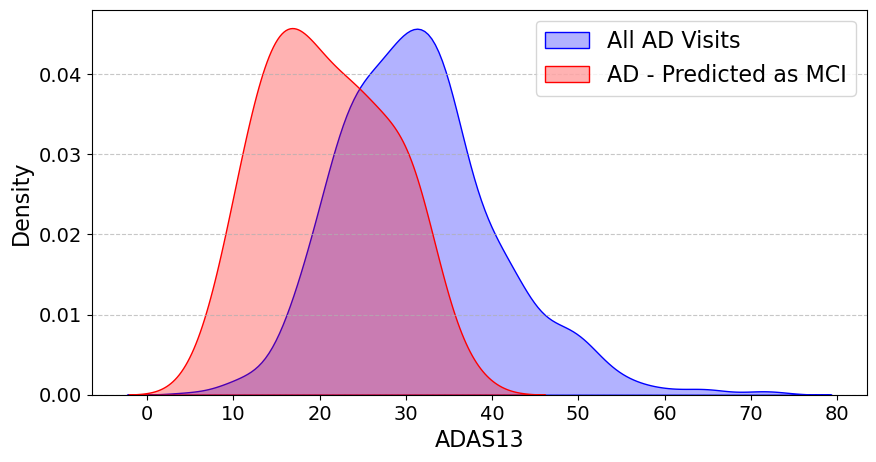}}
\end{figure}

Overall, these findings highlight the importance of model selection in predicting
disease progression. A model that performs well on the majority class but fails to
detect conversion cases may not be useful in certain clinical applications like
Alzheimer's disease prediction. Visit sequence analysis further emphasizes that
Transformer-based models may provide a more useful approach by prioritizing early
detection, even if it comes at the cost of slightly lower performance on stable
sequences.
\section{Limitations and Future Work}
\label{sec:limitation_future_work}

Despite the improvements demonstrated by the Transformer model, several limitations remain. One key limitation is feature selection, as the model was trained on structured clinical data but did not incorporate raw imaging modalities such as MRI or PET scans. While volumetric measures (e.g., hippocampus volume, whole brain volume) and PET-derived biomarkers (e.g., amyloid and tau levels) were included, these features provide summarized information rather than the full spatial detail available in raw scans. Future work could explore the integration of deep learning-based feature extraction from imaging data to enhance the model’s predictive capabilities.

Another notable finding was that the model did not show performance gains when provided with longer visit histories, contrary to expectations. Future research could explore alternative methods to selectively prioritize the most relevant historical data.

Furthermore, conversion prediction remains a difficult task despite the improvements seen with the Transformer model. While the model significantly outperformed RNNs in identifying converters, particularly in shorter visit sequences, performance remains suboptimal, highlighting the inherent challenge of predicting disease transition. By addressing these limitations, future research can refine Transformer-based models for longitudinal disease prediction, ultimately improving their clinical utility in early detection and intervention strategies.

\section{Conclusion}
\label{sec:conclusion}

Our study demonstrates the effectiveness of our proposed Transformer model in predicting Alzheimer's disease progression, particularly in identifying converter sequences. Compared to recurrent neural networks, the Transformer demonstrated a more balanced classification, reducing bias toward the majority class while improving sensitivity to early-stage transitions. The model's improved performance, especially for sequences with shorter visit histories, suggests that attention-based architectures are better suited for early detection tasks.

These findings emphasize the role of model selection in clinical applications where early intervention is critical. By prioritizing the detection of high-risk individuals, Transformer-based models offer a promising approach to improve Alzheimer's disease diagnosis and support timely medical decision-making.

\bibliography{chil-sample}

\end{document}